%% file: main.tex
\documentclass[10pt,twocolumn,letterpaper]{article}

\usepackage[pagenumbers]{cvpr} 

\input{preamble}
\usepackage[table]{xcolor}
\definecolor{cvprblue}{rgb}{0.21,0.49,0.74}
\usepackage[pagebackref,breaklinks,colorlinks,allcolors=cvprblue]{hyperref}

\def\paperID{15275} 
\def\confName{CVPR}
\def\confYear{2026}

\title{PROPA: Toward Process-level Optimization in Visual Reasoning via Reinforcement Learning}

\author{Yanbei Jiang\\
University of Melbourne\\
{\tt\small yanbeij@student.unimelb.edu.au}
\and
Chao Lei\\
University of Melbourne\\
{\tt\small clei1@student.unimelb.edu.au}
\and
Yihao Ding\\
University of Melbourne\\
{\tt\small yihao.ding@uwa.edu.au}
\and
Krista Ehinger\\
University of Melbourne\\
{\tt\small kehinger@unimelb.edu.au}
\and
Jey Han Lau\\
University of Melbourne\\
{\tt\small jeyhan.lau@unimelb.edu.au}
}
\begin{document}
\maketitle
\input{sec/0_abstract}    
\input{sec/1_intro}
\input{sec/2_related}

\input{sec/3_method}
\input{sec/4_experiment}
\input{sec/5_ablation}
\input{sec/6_case_study}

\input{sec/7_conclusion}
{
    \small
\bibliographystyle{ieeenat_fullname}
    \bibliography{main}
}

\input{sec/X_suppl}

\end{document}

%% file: preamble.tex
\usepackage[vlined, ruled]{algorithm2e}
\usepackage{algorithmic}
\usepackage{multirow}
\usepackage{tcolorbox}
\usepackage{tabularx} 
\usepackage{subcaption}
\usepackage{booktabs}
\usepackage{wrapfig}
\usepackage{pifont}
\usepackage{float}
\usepackage{longtable}
\usepackage[table]{xcolor}
\newtcolorbox{promptbox}[1][]{
  colback=gray!5!white,
  colframe=black!75!white,
  fonttitle=\bfseries,
  title=Geometry (Qwen2.5-VL-3B),
  #1
}
\usepackage{xcolor} 
\definecolor{softred}{RGB}{220,80,80}
\definecolor{softgreen}{RGB}{60,160,90}



%% file: sec/0_abstract.tex
\begin{abstract}
Despite significant progress, Vision-Language Models (VLMs) still struggle with complex visual reasoning, where multi-step dependencies cause early errors to cascade through the reasoning chain. Existing post-training paradigms are limited: Supervised Fine-Tuning (SFT) relies on costly step-level annotations, while Reinforcement Learning with Verifiable Rewards (RLVR) methods like GRPO provide only sparse, outcome-level feedback, hindering stable optimization.
We introduce PROPA — \textbf{P}rocess-level \textit{R}easoning \textbf{O}ptimization with interleaved \textbf{P}olicy \textbf{A}lignment, a novel framework that integrates Monte Carlo Tree Search (MCTS) with GRPO to generate dense, process-level rewards and optimize reasoning at each intermediate step without human annotations. To overcome the cold-start problem, PROPA interleaves GRPO updates with SFT, enabling the model to learn from both successful and failed reasoning trajectories. A Process Reward Model (PRM) is further trained to guide inference-time search, aligning the test-time search with the training signal. Across seven benchmarks and four VLM backbones, PROPA consistently outperforms both SFT- and RLVR-based baselines. It achieves up to 17.0\% gains on in-domain tasks and 21.0\% gains on out-of-domain tasks compared to existing state-of-the-art, establishing a strong reasoning and generalization capability for visual reasoning tasks. The code is available at: \href{https://github.com/YanbeiJiang/PROPA}{https://github.com/YanbeiJiang/PROPA}.
\end{abstract}

%% file: sec/1_intro.tex
\vspace{-0.3cm}
\section{Introduction}
\label{sec:intro}

\begin{figure}[t]
  \centering
  \includegraphics[width=\linewidth]{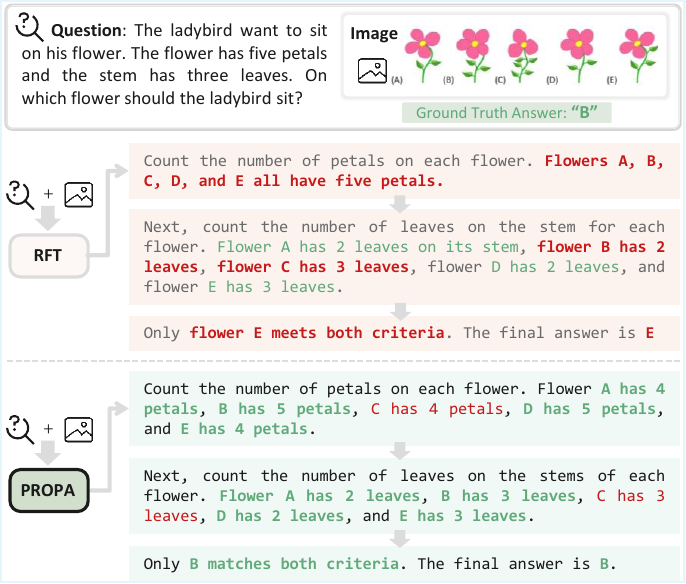}
  \caption{Example outputs of our PROPA framework compared with RFT (Reinforcement Fine-Tuning using GRPO)}
  \label{fig:intro}
  \vspace{-0.5cm}
\end{figure}

Vision-Language Models (VLMs) excel at a wide range of visual-text tasks such as visual question answering~\cite{goyal2017making,marino2019ok,ding2023vqa} and image captioning~\cite{saito2023pic2word,vinyals2015show}. However, as the field shifts toward more challenging reasoning-oriented tasks, the limitations of current VLMs become more evident. Unlike conventional multimodal tasks that focus on recognition or description, visual reasoning requires a model to understand complex relationships, perform comparisons, and draw inferences through a structured reasoning process, as shown in example of Figure \ref{fig:intro}. Such tasks often involve multi-step problem solving, where earlier reasoning steps serve as a foundation for subsequent conclusions~\cite{jiang2025beyond,liu2025small}. Errors made at intermediate steps can easily propagate such as miscounting the number of petals, leading to incorrect final answers. This stepwise dependency makes visual reasoning inherently more difficult, and despite recent advances, VLMs still struggle to produce consistent and reliable results on such tasks~\cite{xu2025visulogic,jiang2024marvel,kim2025enhancing}.

Advances in reasoning models such as OpenAI-o1~\cite{o1} and Kimi-K1.5~\cite{team2025kimi} have sparked increasing interest in extending structured long Chain-of-Thought (CoT) reasoning to VLMs. Existing post-training approaches can be broadly categorized into two paradigms:
(i) Supervised Fine-Tuning (SFT)-based methods, which train over long-form reasoning by leveraging labeled reasoning chains or synthetically generated intermediate data~\cite{luo2025ursa,yang2025re}; and
(ii) Reinforcement Learning with Verifiable Rewards (RLVR) methods such as Group Relative Policy Optimization (GRPO)~\cite{guo2025deepseek}, which have become increasingly popular~\cite{tan2025reason, liu2025visual,peng2025lmm,huang2025vision}. These approaches demonstrate stronger reasoning capabilities and better generalization by rewarding trajectories containing correct answers.

Despite recent progress, both paradigms suffer from notable limitations. First, SFT-based post-training suffers from error propagation issue, as inaccuracies in the generated intermediate reasoning steps can easily accumulate and lead to incorrect final answers. This issue is particularly severe when step-level annotations are automatically generated rather than manually verified~\cite{zhang2024restmcts,yang2025re,yao2024mulberry,zhang2024improve,wu2025sdrt}.
Moreover, SFT tends to overfit to specific training trajectories, reducing its generalization to diverse reasoning paths~\cite{li2025perception}.
Second, RLVR-based methods face challenges in exploration depth and training stability. 
Since they rely solely on outcome-based rewards~\cite{tan2025reason, liu2025visual}, they often suffer from reward sparsity, only a small portion of model-generated reasoning paths receive high positive rewards, leading to unstable training.

To address these challenges, we introduce a novel framework, PROPA (Process-level Reasoning Optimization with interleaved Policy Alignment), in which the policy model is optimized using the GRPO, where the reasoning chains are generated under the Monte Carlo Tree Search (MCTS) paradigm. The key idea is to leverage the exploratory capacity of MCTS to navigate the vast search space of possible reasoning paths. Through its intrinsic exploration and reward backpropagation mechanisms, MCTS naturally provides dense, process-level reward signals, thereby removing the need for explicit human annotations. 
Next, we apply a localized GRPO update using the MCTS-derived rewards to optimize the policy at each intermediate reasoning step, rather than only at the final answer. Our goal is to encourage the model to generate intermediate trajectories that are more likely to lead to correct final answers.
Our methodology also introduces several key innovations to ensure robust and effective training. First, relying solely on MCTS-guided GRPO suffers from a severe cold-start problem, where an untrained policy often leads the search toward uninformative paths, preventing reward signals and halting learning. To address this, we propose an Interleaved GRPO and SFT scheme, where MCTS-guided GRPO is applied to successful cases for refinement, while SFT is used on failed cases to preserve basic capabilities and prevent catastrophic forgetting. Moreover, to align the test-time search with the training signal, we introduce a Process Reward Model (PRM) that approximates process rewards discovered during training, serving as a heuristic to guide MCTS exploration at inference. 

In summary, our work makes the following contributions: 1) We propose a novel MCTS-guided GRPO framework that automatically generates and utilizes process-level rewards for visual reasoning, circumventing the need for dense, manual step-by-step annotations. 2) We introduce an interleaved GRPO and SFT training strategy that effectively addresses the cold-start problem in RL-based training and enables the model to learn robustly from both successful and unsuccessful reasoning attempts. 3) We develop a PRM that serves as a heuristic to guide MCTS during test time, aligning the search process with the training signal and significantly improving the accuracy of the final answer.

%% file: sec/2_related.tex
\section{Related Work}
\label{sec:formatting}
\vspace{-0.1cm}
\paragraph{Visual Reasoning} Visual reasoning requires models to perform multi-step cognitive processes grounded in visual inputs, enabling sequential inference over compositional, mathematical, geometric, and scientific concepts~\cite{lu2024mathvista,zou2024dynamath,gao2023g}.
Some prior efforts rely on SFT to enhance reasoning performance. For instance, LLaVA-CoT~\cite{xu2024llava} employs a multi-stage SFT framework incorporating CoT annotations to strengthen step-wise reasoning.
Recent studies further explore automatic generation of intermediate reasoning steps by leveraging tree search–based strategies. The work~\cite{yang2025re} enriches reasoning traces through a context-augmented knowledge base and re-ranking via tree, while Mulberry~\cite{yang2025re} constructs new reasoning datasets through MCTS-guided data synthesis for subsequent SFT.
Other approaches adopt distillation-based frameworks to extract and refine reasoning traces. SDRT~\cite{wu2025sdrt}, for example, introduces a self-distillation method that diversifies and deepens reasoning trajectories learned from the model itself. Moreover, A*Star~\cite{wu2025boosting} applies MCTS during inference to dynamically explore and execute optimal reasoning paths. These approaches remain inherently limited by the reliability of auto-generated traces, often resulting in noisy supervision. 

\vspace{-0.1cm}
\paragraph{Reinforcement learning}
Reinforcement learning (RL) has emerged as a powerful paradigm for improving the reasoning ability of VLMs. Early efforts, such as Reinforcement Learning from Human Feedback (RLHF)~\cite{ouyang2022training} and Direct Preference Optimization (DPO)~\cite{rafailov2023direct}, primarily focused on aligning model outputs with human preferences. More recent developments such as GRPO~\cite{xu2025deepseek}, which employs rule-based scoring. Building upon these foundations, several works have incorporated RL within multimodal alignment frameworks. For example, Insight-V~\cite{dong2024insight} integrates SFT and DPO through a two-agent training pipeline, while Reason-RFT~\cite{tan2025reason}, Visual-RFT~\cite{liu2025visual}, and R1-V~\cite{chen2025r1v} extend the principles of RLVR from DeepSeek-R1 to visual reasoning tasks such as object detection and counting. More recently, MM-Eureka~\cite{meng2025mm} and MMR1~\cite{leng2025mmr1} further explored RLVR for multimodal mathematical reasoning, achieving strong generalization abilities. Despite these advances, most existing methods remain outcome-based. Our work addresses this gap by proposing a mechanism to generate robust process-level rewards in the visual domain, overcoming the signal inaccuracy that limits existing approaches.

%% file: sec/3_method.tex
\section{Methods}
\begin{figure*}[t]
  \centering
  \includegraphics[width=0.95\linewidth]{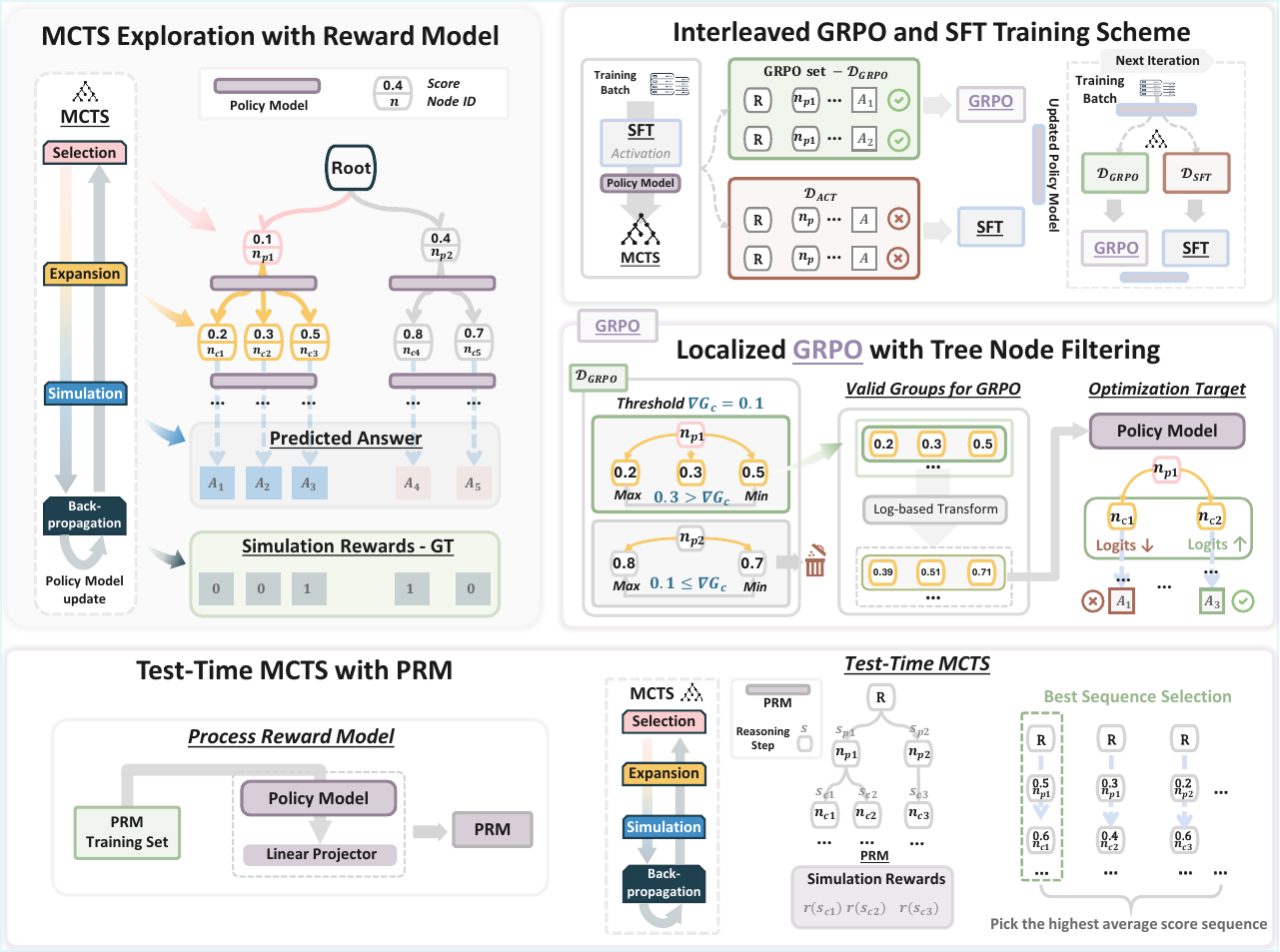}
  \vspace{-0.2cm}
  \caption{Overview of the proposed PROPA framework. The architecture integrates MCTS-guided process-level reward generation, an interleaved GRPO training scheme, and a learned PRM for test-time inference.}
  \label{fig:framework}
  \vspace{-0.4cm}
\end{figure*}

Our proposed PROPA framework enhances the visual reasoning capabilities of VLMs through a novel reinforcement learning approach guided by MCTS. The methodology is characterized by three core components, as illustrated in Figure \ref{fig:framework}: (1)  MCTS-based process-level rewards generation to guide the GRPO training process;
(2) an interleaved training scheme to balance exploration and exploitation; and (3) a PRM to enable efficient and robust test-time inference.

\subsection{Problem Formulation}
Given an image input $I$ and a corresponding textual question $Q$, the task of a visual reasoning model is to generate a sequence of reasoning steps $\rho = (s_0,\ldots, s_n)$ that culminates in a correct final answer $\mathcal{A}$, where $s \in \rho$,  denotes a reasoning step.  The initial state $s_0$ encodes the $I$ and $Q$. $s_n$ is a terminal state, denoted  as $s_t$, when it produces the final answer $\mathcal{A}$.  The reasoning process $\rho$ is also known as the Chain-of-Thought (CoT). Our goal is to learn a policy $\pi_{\theta}$, parameterized by $\theta$, which generates the reasoning sequence step-by-step. The policy model takes the current reasoning sequence $\rho=\{ s_0, \ldots, s_{n-1}\}$ and produces the next state $s_n$. The objective is to maximize the expected reward for each state in $\rho$, where the reward is determined by the MCTS-based simulation introduced later.

\subsection{MCTS-Guided GRPO Training}

\subsubsection{MCTS for Process Reward Generation}
For each training instance, we construct a search tree in which each node is defined as a set: 
\begin{equation}
n = \{s, n_p, \mathcal{W}(s), \mathcal{N}(s)\},
\end{equation}
where state $s$ denotes a reasoning step in $\rho$, $n_p$ is its parent node (with the relation represented as $n_p \!\leadsto\! n$), $\mathcal{W}(s)$ represents the total accumulated simulation reward, $\mathcal{N}(s)$ denotes the number of visits to the state $s$, and $n_0$ represents the root node. 
A node $n=n_t$ is considered as a terminal node if its state $s = s_t, s \in n$ contains the final answer $\mathcal{A}$. We note that each node $n$ is uniquely determined by its state $s$, as a results, we use the same notation $\leadsto$ to represents the ancestry relation for state. The overall MCTS procedure iterates through four stages:

\begin{enumerate}
    \item \textbf{Selection:} Starting from the root, we traverse the tree by recursively selecting the node with the highest Upper Confidence Bound (UCT) for score  $v(s)$ computed as:

\begin{equation}
v(s) = \mathcal{Q}(s) + \mathcal{C} \cdot \sqrt{\frac{\ln \mathcal{N}(s_p)}{\mathcal{N}(s)}},
\end{equation}

where $\mathcal{Q}(s) = \frac{\mathcal{W}(s)}{\mathcal{N}(s)}$ is the average simulation reward of state $s$. We note that $v(s)=\infty$ when $\mathcal{N}(s)=0$ and $\mathcal{C}$ is an exploration constant.
    

    \item \textbf{Expansion:} From the selected  node $n_z$, we expand $k$ child nodes $\{ n_{c_1}, \ldots, n_{c_k} \mid n_{z}\!\leadsto\! n_{c_i},\, i = 1, \ldots, k \}$
to sample the next reasoning steps $k$ times from the current policy $\pi_{\theta}$, given  all reasoning steps $\rho=\{s_0 \!\leadsto\! \cdots \!\leadsto\! s_z\}$. 

    \item \textbf{Simulation:} For each newly expanded node $n_{c_i}$, a Monte Carlo rollout is performed by sampling subsequent reasoning paths from state $s_{c_i}$ until a final answer $\mathcal{A}$ is reached. The simulation then returns a simulation reward $r(s_{c_i})$, where $r(s_{c_i}) = 1$ if the final answer $\mathcal{A}$ is correct, and $r(s_{c_i}) = 0$ otherwise.

    \item \textbf{Backpropagation:} The simulation reward is backpropagated along the path 
$\rho=\{n_{0} \!\leadsto\! \cdots \!\leadsto\! n_{c_i}\}$, 
where the statistics of each node on this path are updated as:
\begin{equation}
\mathcal{N}(s) \leftarrow \mathcal{N}(s) + 1, \quad
\mathcal{W}(s) \leftarrow \mathcal{W}(s) + r(s_{c_i}),
\end{equation}

\end{enumerate}
After multiple iterations, the value $\mathcal{Q}(s)$ of a node serves as dense, process-level reward estimates.

\subsubsection{Localized GRPO with Tree Node Filtering}
\label{grpo}


For each MCTS-expanded tree, we construct the training data for GRPO by traversing all expansion steps from top to bottom. For each parent node $n_z$, its corresponding reasoning chain $\rho = \{ s_0 \leadsto \cdots  \!\leadsto\! s_z \}$ serves as the given problem in GRPO, while the expanded child nodes constitute a group $G_c = \{n_{c_1}, \ldots, n_{c_k} \mid n_z \!\leadsto\! n_{c_i},\, i = 1, \ldots, k \}$ of GRPO completions. The GRPO reward of each child node $n_{c_i} \in G_c$, used for advantage computation, is given by its $\mathcal{Q}(s_{c_i})$ value, which is obtained during the full MCTS process. Formally, the advantage of GRPO loss for $i$-th child is given by:
\begin{equation}
A_i = \frac{\mathcal{Q}(s_{c_i}) - \operatorname{mean}(\{\mathcal{Q}(s_{c_i}), i = 1, \ldots, k \})}
           {\operatorname{std}(\{\mathcal{Q}(s_{c_i}), i = 1, \ldots, k \})}.
\end{equation}


As GRPO learns from the advantage difference within each group, to ensure meaningful learning signals, we perform  group-level filtering and transformation operations. If all child nodes in a group have very similar values, the signal becomes weak. To address this, we first apply the child node filtering: for a group $G_c$, we calculate the value difference of expanded children nodes  $\Delta_{G_c} = \max(\mathcal{Q}(s_{c_i})) - \min(\mathcal{Q}(s_{c_i}))$ and apply this to each group. If $\Delta_{G_c}$ is below a predefined threshold $\tau$, we discard this group. We then apply a log-based nonlinear transformation to enhance the contrast among values:
\begin{equation}
\small
\mathcal{Q}(s_{c_i}) = 
\operatorname{clip}\!\left(
\dfrac{\log(1 + \alpha \mathcal{Q}(s_{c_i}))}{\log(1 + \alpha)},\, 0,\, 1
\right)
\quad i = 1, \ldots, k,
\end{equation}
where $\alpha$ is a scaling parameter and $\operatorname{clip}$ represents clipping. This transformation amplifies larger values while keeping smaller ones suppressed, effectively increasing the variance among child node values.
Through this process, the policy is optimized toward intermediate reasoning steps that contribute to correct final answers.

\subsection{Interleaved GRPO and SFT Training Scheme}
Following most prior works~\cite{zhang2024restmcts,tan2025reason,guo2025deepseek,liu2025visual}, we first perform a SFT activation stage based on CoT annotations. Given the training dataset $\mathcal{D}$, we use a larger teacher model (GPT-4.1~\cite{gpt4o}) to produce sequences of reasoning steps $\rho = \{s_0, s_1,\ldots, s_n\}$ given each instance and answer in $\mathcal{D}$, resulting in the activation dataset $\mathcal{D}_\text{act}$. We leverage $\mathcal{D}_\text{act}$ to teach the model to generate the step-by-step reasoning traces structure and appropriate termination tokens. 
Then, to alleviate the cold-start issue of GRPO, we propose an interleaved training scheme that dynamically partitions training data according to the MCTS outcomes. For training instances $\{x_1, x_2, \dots, x_n\}$ in $\mathcal{D}$, each instance $x_j$, $j=1,\ldots,n$, is used to construct a search tree $\mathcal{T}_j$. Let $N_{t_j} = \{n_t \in \mathcal{T}_j\}$ denote the set of terminal nodes in $\mathcal{T}_j$. We generate the training data and divide it into two non-overlapping subsets:
\vspace{-0.2cm}
\begin{align}
\small
\mathcal{D}_{\text{grpo}} &= \{  (\rho, G) \in \mathcal{T}_j \mid \exists s_t \in n_t \in N_{t_j}, r(s_t) = 1\}, \\[4pt]
\mathcal{D}_{\text{sft}}  &= \{j \mid \forall s_t \in n_t \in N_{t_j}, r(s_t) = 0\},
\end{align}
where $r(s_t)$ is the simulation reward for $s_t$ determined during MCTS construction. $\mathcal{D}_{\text{grpo}}$ comprises all input–completion pairs, $(\rho, G)$, for GRPO, generated from trees where MCTS successfully finds at least one correct terminal node, while $\mathcal{D}_{\text{sft}}$ contains the indexes of those where all simulations fail.

Then, the $\mathcal{D}_{\text{grpo}}$ is used for parameter updates with GRPO (as mentioned in Section \ref{grpo}). The $\mathcal{D}_{\text{sft}}$ is used for locating corresponding $j$-th instances in the activation dataset $\mathcal{D}_\text{{act}}$ to optimize with SFT cross-entropy loss. To ensure stable training,
both GRPO and SFT updates are executed periodically every $\lambda$ iterations, once sufficient training data have been collected. Across training iterations, GRPO and SFT are interleaved to continuously refine valid reasoning trajectories while recovering knowledge from failed cases. 



\subsection{Test-Time MCTS with a PRM}
Traditional test-time algorithms such as greedy search or best-N search~\cite{brown2024large} can only explore a limited portion of the search tree, while performing a full simulation to obtain terminal rewards during inference is computationally infeasible. To enable efficient and effective exploration of the search space, we introduce a test-time MCTS mechanism. Specifically, we train a Process Reward Model (PRM), denoted as $P_{\phi}$, which extends the initial policy model $\pi_{\theta}$ by adding a linear layer after its output layer to directly predict node scores.
The PRM is trained using data collected from all successful MCTS search trees generated during the final epoch of the interleaved training phase and optimized using Mean Square Error (MSE) loss to all parameters.

During test-time MCTS, the PRM replaces the traditional simulation step. When a new node is expanded, instead of performing a simulation step, we query the PRM to directly estimate its value. Specifically, given the parent node $n_z$, for each newly expanded node $n_{c_i}$, the PRM takes $\{s_0 \!\leadsto\! \cdots \!\leadsto\! s_{c_i} \}$ as input, predicts its simulation rewards $r(s_{c_i}) = \operatorname{clip}(P_{\phi}(s_{c_i}), 0, 1)$ as the target. The predicted reward is then backpropagated up the tree to update all parent nodes, following the standard MCTS backpropagation procedure. 
This design aligns the test-time search with the training signal, leveraging both the exploration and exploitation characteristics of MCTS. 

Once the maximum iteration count is reached, we identify the set of all terminal nodes $N_t$ within the search tree that represent a complete answer. For each terminal node $n_t \in N_t$, we trace its reasoning path $ \rho=\{s_{0} \!\leadsto\! \cdots \!\leadsto\! s_t\}$ and compute the average $\mathcal{Q}(s)$ value of nodes along this path. The path with the highest average $\mathcal{Q}(s)$ value, denoted as $\rho^*$, is selected, and the answer $\mathcal{A}$ contained in the terminal node $n_t \in \rho^*$ serves as the final answer. Specifically, the selected terminal node $n_t$ is determined by:

\begin{equation}
n_t = \underset{n_t \in N_t}{\arg\max} \left( \frac{1}{|\rho|} \sum_{s \in n \in \rho} \mathcal{Q}(s) \right)
\end{equation}
Algorithm \ref{alg:mcts_grpo_training_compact} provides an overview of the overall workflow and sequential process of our PROPA framework.

\begin{algorithm}[ht]
\caption{PROPA Framework}
\renewcommand{\algorithmicrequire}{\textbf{Input:}}
\renewcommand{\algorithmicensure}{\textbf{Output:}}
\label{alg:mcts_grpo_training_compact}
\begin{algorithmic}[1]
\REQUIRE Base model $\pi_\theta$, training dataset $\mathcal{D}$, testing $\mathcal{D}_\text{test}$, GRPO dataset $\mathcal{D}_\text{grpo}$, SFT dataset $\mathcal{D}_\text{sft}$, \\teacher model $\pi_{\text{teacher}}$, training epochs $E$.

\STATE $\mathcal{D}_{\text{act}} \leftarrow \text{GenerateCoT}(\pi_{\text{teacher}}, \mathcal{D})$
\STATE $\pi_\theta \leftarrow \text{SFT}(\pi_\theta, \mathcal{D}_{\text{act}})$ \tcp{SFT activation}
\STATE $\mathcal{D}_{\text{prm}} \leftarrow \emptyset$, $\mathcal{D}_{\text{grpo}} \leftarrow \emptyset$,$\mathcal{D}_{\text{sft}} \leftarrow \emptyset$

\STATE \textit{\textbf{Interleaved Training Loop}}
\FOR {$e = 1$ to $E$}
     
    \FOR {$x$ in $\mathcal{D}$}
        \STATE $\mathcal{T} \leftarrow \text{build\_MCTS}(\pi_\theta, x)$
                
        \STATE $\mathcal{D}_{\text{grpo}} \leftarrow \mathcal{D}_{\text{grpo}} \cup\text{partition\_GRPO}(\mathcal{T})$
        \STATE $\mathcal{D}_{\text{sft}}\leftarrow \mathcal{D}_{\text{sft}} \cup \text{partition\_SFT}(\mathcal{T})$
        \STATE 
        $\mathcal{D}_{\text{grpo}} \leftarrow \text{filter\_and\_transform}(\mathcal{D}_{\text{grpo}})$

        \IF {meet the iteration $\lambda$}
        \STATE $\pi_\theta \leftarrow \text{localized\_GRPO}(\pi_\theta, \mathcal{D}_{\text{grpo}})$
        \STATE $\pi_\theta \leftarrow \text{SFT}(\pi_\theta, \mathcal{D}_{\text{sft}})$
       \STATE $\mathcal{D}_{\text{grpo}} \leftarrow \emptyset$, $\mathcal{D}_{\text{sft}} \leftarrow \emptyset$
        \ENDIF
        \IF {$e=E$}
                \STATE $\mathcal{D}_{\text{prm}} \leftarrow \mathcal{D}_{\text{prm}} \cup \bigcup_{s \in \mathcal{T}} \{(s, \mathcal{Q}(s))\}$
        \ENDIF
    \ENDFOR
\ENDFOR

\STATE \textit{\textbf{PRM Training \& Test-Time Usage}}
\STATE $\phi \leftarrow \text{train\_PRM}(\mathcal{D}_{\text{PRM}})$
\STATE $\mathcal{T}_{test} \leftarrow \text{build\_MCTS\_with\_PRM}(\pi_\theta, \mathcal{D}_{test}, \phi)$ 
\STATE $n_t \leftarrow \text{highest\_average\_value}(\mathcal{T}_{test})$ 

\end{algorithmic}
\end{algorithm}
\vspace{-0.2cm}

%% file: sec/4_experiment.tex
\section{Experiments Setup}

\definecolor{LightBlue}{RGB}{232,243,255}
\definecolor{LightGray}{RGB}{240,240,240}
\begin{table*}[ht]
\centering
\caption{Performance comparison across \textbf{Structure Perception}, \textbf{Spatial Transformation}, and \textbf{Math/Science} domains on ID and OOD datasets. Each test is run three times, and the reported results are the averaged scores. The best results are in bold. A dash (–) indicates that the output does not conform to the required format.}
\resizebox{\textwidth}{!}{
\begin{tabular}{lccccccc}
\toprule
\multirow{2}{*}{\textbf{Model \& Method}} &
\multicolumn{2}{c}{\textbf{Structure Perception}} & 
\multicolumn{3}{c}{\textbf{Spatial Transformation}} &
\multicolumn{2}{c}{\textbf{Math/Science}} \\
\cmidrule(lr){2-3} \cmidrule(lr){4-6} \cmidrule(lr){7-8}
 & GeoMath (ID) & Geometry3K (OOD) & Trance (ID) & TranceL (OOD) & TranceR (OOD) & MathVision (ID) & DynaMath (OOD) \\
\midrule
\rowcolor{LightGray}
\multicolumn{8}{l}{\textbf{Close-Source Models}} \\
\midrule
GPT-4.1~\cite{gpt4o} & 43.29 & 32.25 & - & - & - & 21.71 & 38.68 \\
openai-o3~\cite{o3} & 55.98 & 87.56 & - & - & - & 58.17 & 61.18 \\
\midrule
\rowcolor{LightGray}
\multicolumn{8}{l}{\textbf{Qwen2.5-VL-3B}} \\
\midrule
Zero-Shot & 39.39 & 37.00 & 1.64 & 1.39 & 1.57 & 18.09 & 35.10 \\
SFT-F & 43.05 & 40.25 & 8.98 & 7.81 & 7.67 & 23.36 & 36.68 \\
SFT-CoT & 50.12 & 46.88 & 47.60 & 32.96 & 32.96 & 34.87 & 35.53 \\
RFT-zero & 42.20 & 43.12 & 2.83 & 2.13 & 2.66 & 16.45 & 36.68 \\
RFT & 48.90 & 47.12 & 46.69 & 35.38 & 37.22 & 29.93 & 34.24 \\
\textbf{Ours} & \cellcolor{LightBlue}\textbf{55.00} & \cellcolor{LightBlue}\textbf{54.37} & \cellcolor{LightBlue}\textbf{56.32} & \cellcolor{LightBlue}\textbf{47.59} & \cellcolor{LightBlue}\textbf{45.70} & \cellcolor{LightBlue}\textbf{38.49} & \cellcolor{LightBlue}\textbf{37.54} \\
\midrule
\rowcolor{LightGray}
\multicolumn{8}{l}{\textbf{Qwen2.5-VL-7B}} \\
\midrule
Zero-Shot & 43.54 & 45.50 & 15.13 & 9.96 & 10.03 & 24.01 & 37.54 \\
SFT-F & 48.66 & 47.88 & 28.05 & 21.95 & 20.80 & 23.03 & 39.26 \\
SFT-CoT & 53.05 & 55.75 & 55.88 & 36.95 & 35.63 & 33.55 & 40.26 \\
RFT-zero & 51.34 & 54.12 & 7.31 & 4.73 & 4.06 & 24.01 & 42.41 \\
RFT & 53.17 & 54.75 & 60.44 & 46.72 & 44.01 & 30.26 & 43.41 \\
\textbf{Ours} & \cellcolor{LightBlue}\textbf{56.10} & \cellcolor{LightBlue}\textbf{57.38} & \cellcolor{LightBlue}\textbf{62.82} & \cellcolor{LightBlue}\textbf{50.03} &\cellcolor{LightBlue} \textbf{48.60} & \cellcolor{LightBlue}\textbf{44.74} & \cellcolor{LightBlue}\textbf{43.55} \\
\midrule
\rowcolor{LightGray}
\multicolumn{8}{l}{\textbf{Intern2.5-VL-2B}} \\
\midrule
Zero-Shot & 17.07 & 36.62 & - & - & - & 6.58 & 10.46 \\
SFT-F & 41.71 & 51.50 & 13.73 & 11.10 & 12.79 & 19.74 & 28.08 \\
SFT-CoT & 42.93 & 49.12 & 54.46 & 43.06 & 43.77 & \textbf{36.51} & 28.37 \\
RFT-zero & 24.88 & 23.50 & 0.00 & 0.00 & 0.00 & 9.87 & 20.20 \\
RFT & 45.49 & 45.88 & 42.18 & 26.33 & 26.48 & 29.93 & 28.94 \\
\textbf{Ours} & \cellcolor{LightBlue}\textbf{47.68} & \cellcolor{LightBlue}\textbf{52.88} & \cellcolor{LightBlue}\textbf{59.13} & \cellcolor{LightBlue}\textbf{47.37} & \cellcolor{LightBlue}\textbf{46.69} & \cellcolor{LightBlue}31.88 & \cellcolor{LightBlue}\textbf{30.66} \\
\midrule
\rowcolor{LightGray}
\multicolumn{8}{l}{\textbf{Intern2.5-VL-8B}} \\
\midrule
Zero-Shot & 17.07 & 36.25 & 5.65 & 4.62 & 4.47 & 17.76 & 17.77 \\
SFT-F & 50.00 & 59.75 & 39.64 & 33.85 & 35.71 & 25.66 & 34.81 \\
SFT-CoT & 55.61 & 56.88 & 71.17 & 57.76 & 56.53 & \textbf{39.14} & 38.68 \\
RFT-zero & 45.98 & 58.75 & 1.60 & 1.89 & 1.56 & 19.74 & 33.95 \\
RFT & 52.07 & 60.88 & 69.37 & 55.19 & 55.39 & 36.84 & 37.68 \\
\textbf{Ours} & \cellcolor{LightBlue}\textbf{58.41} & \cellcolor{LightBlue}\textbf{61.00} & \cellcolor{LightBlue}\textbf{74.33} & \cellcolor{LightBlue}\textbf{62.53} & \cellcolor{LightBlue}\textbf{61.69} & \cellcolor{LightBlue}38.16 & \cellcolor{LightBlue}\textbf{40.10} \\
\bottomrule
\end{tabular}
}
\label{tab:main_results}
\end{table*}

\paragraph{Datasets.} 
To evaluate our method, we conduct experiments on seven benchmark datasets across three challenging domains of visual reasoning, each comprising an in-domain (ID) and a corresponding out-of-domain (OOD) set. All models, including baselines, are trained and validated on the ID datasets and evaluated on all seven test sets.
(1) Geometric \textbf{Structure Perception} focuses on analyzing geometric relationships and spatial layouts, with Geo170K~\cite{geo170k} and Math360K~\cite{math360k} as ID dataset, GeoMath, and Geometry3K as the OOD benchmark.
(2) Visual \textbf{Spatial Transformation} examines 3D spatial transformation understanding, trained on Trance~\cite{hong2021transformation} and tested on TranceL/R to assess generalization across unseen viewpoints. 
(3) Mathematical and Scientific Reasoning (\textbf{Math/Science}) assesses quantitative problem solving grounded in visual information, using MathVision~\cite{wang2024measuring} for ID training and DynaMath~\cite{zou2024dynamath} for OOD evaluation.
Examples from each dataset, comparisons between ID and OOD cases, along with the statistics of the train, validation, and test splits, are provided in Appendix~\ref{sec:datasets}.

\vspace{-0.3cm}
\paragraph{Evaluation Metrics.} We evaluate the performance of all models using Pass@1 accuracy as the primary metric. The specific criteria for determining correctness vary by the answer format:
For numerical answers, correctness is verified by checking for mathematical equivalence. For multiple-choice questions, we perform a direct string match between the model's selected option and the correct choice. For function-type sequences in Trance datasets, the accuracy is calculated as the ratio of correctly predicted functions to the total number of functions in the ground-truth.
\vspace{-0.3cm}
\paragraph{Baselines.} 
To evaluate our method, we compare it against baselines representing two main paradigms. The first is SFT, which includes SFT-F (trained on final answers only) and SFT-CoT (trained on the full CoT process, i.e., the activation dataset $\mathcal{D}_{\text{SFT}}$ generated by the teacher model). The second is Reinforcement Learning (RL). Several prior studies employ GRPO with rewards derived from the final answer’s correctness and format~\cite{liu2025visual,tan2025reason,chen2025r1v}. Among them, we adopt the well recognized Reason-RFT framework~\cite{tan2025reason}, which includes two variants: Reason-RFT-zero (RFT-zero), a pure GRPO-based approach, and Reason-RFT (RFT), which incorporates an initial SFT-CoT activation stage prior to RL optimization. All paradigms were benchmarked across four VLMs: Qwen2.5-VL-Instruct~\cite{yang2024qwen25} and Intern2.5-VL~\cite{chen2024internvl}, each at two different parameter scales. Besides these, we also include two more close-sourced strong baselines, GPT-4.1~\cite{achiam2023gpt} and openai-o3~\cite{o3}.
\vspace{-0.1cm}
\paragraph{Implementation Details.} 
Our implementation is based on the open-source frameworks MS-Swift~\cite{swift} and vLLM~\cite{vllm}. All experiments are conducted on 4$\times$A100 80GB GPUs. The reported results correspond to the best-performing checkpoint, selected according to the highest validation set accuracy, and evaluated on the test set. All training uses LoRA with the rank of 8 and alpha of 32.
To ensure fair comparison, for both baselines and our method, we use a total of 10 epochs, consisting of 3 epochs for the activation stage and 7 epochs for training phase, with a learning rate of 1e-6 and a batch size of 4. During the MCTS process, we promote response diversity by setting the temperature to 1.2, top\_k to 50, top\_p to 0.95, and generating 4 children per node. The training iteration $\lambda$ is set to 40. The number of MCTS iterations is 25, the UCT exploration weight is 1.0, and the maximum simulation depth is 8. For the GRPO stage, we set the data pruning coefficient $\tau$ = 0.1, the logarithmic transformation coefficient to 10, the maximum prompt length is 8192, the KL divergence is 1e-3 and the maximum response length is 1024. The prompts used for both the baselines and our methods, as well as those for generating the activation datasets, are provided in Appendix~\ref{sec:prompts}. Furthermore, Appendix~\ref{sec:time} reports the reference training and inference times for both our method and all baselines.

\vspace{-0.1cm}
\section{Results and Discussion}
\subsection{Main Results}

Table \ref{tab:main_results} compares our PROPA framework against five baseline methods across seven benchmarks evaluated under both ID and OOD conditions when using four VLMs variants of varying parameter scales.  Across all evaluated backbones, our framework mostly achieves the highest accuracy, demonstrating superior structural reasoning and generalization capability. Compared to RFT, the most competitive baseline, ours achieves an average accuracy improvements of 7.9\% and 10.1\% on the Qwen2.5-VL-3B  and Intern2.5-VL-2B backbones, with the largest observed gain of 21.0\% on the TranceL (OOD) benchmark with Intern2.5-VL-2B, demonstrating its enhanced cross-domain generalization capability. When scaled to larger backbones, Qwen2.5-VL-7B and Intern2.5-VL-8B, our method maintains state-of-the-art performance across nearly all benchmarks, outperforming RFT by approximately 4.4\% and 4.1\%, respectively. Furthermore, as the teacher model was given the ground truth answer to generate CoT steps, it is unsurprising that our framework consistently outperform the teacher model through benefiting from the SFT activation stage. However, there remains a noticeable gap compared to OpenAI-o3, which may be attributed either to its larger parameter size or potential data contamination, as suggested by the unexpectedly poor performance on Trance.



\subsection{Performance Over Steps}

\begin{figure}[h]
  \centering

    \vspace{-0.1cm}
    \includegraphics[width=1\columnwidth]{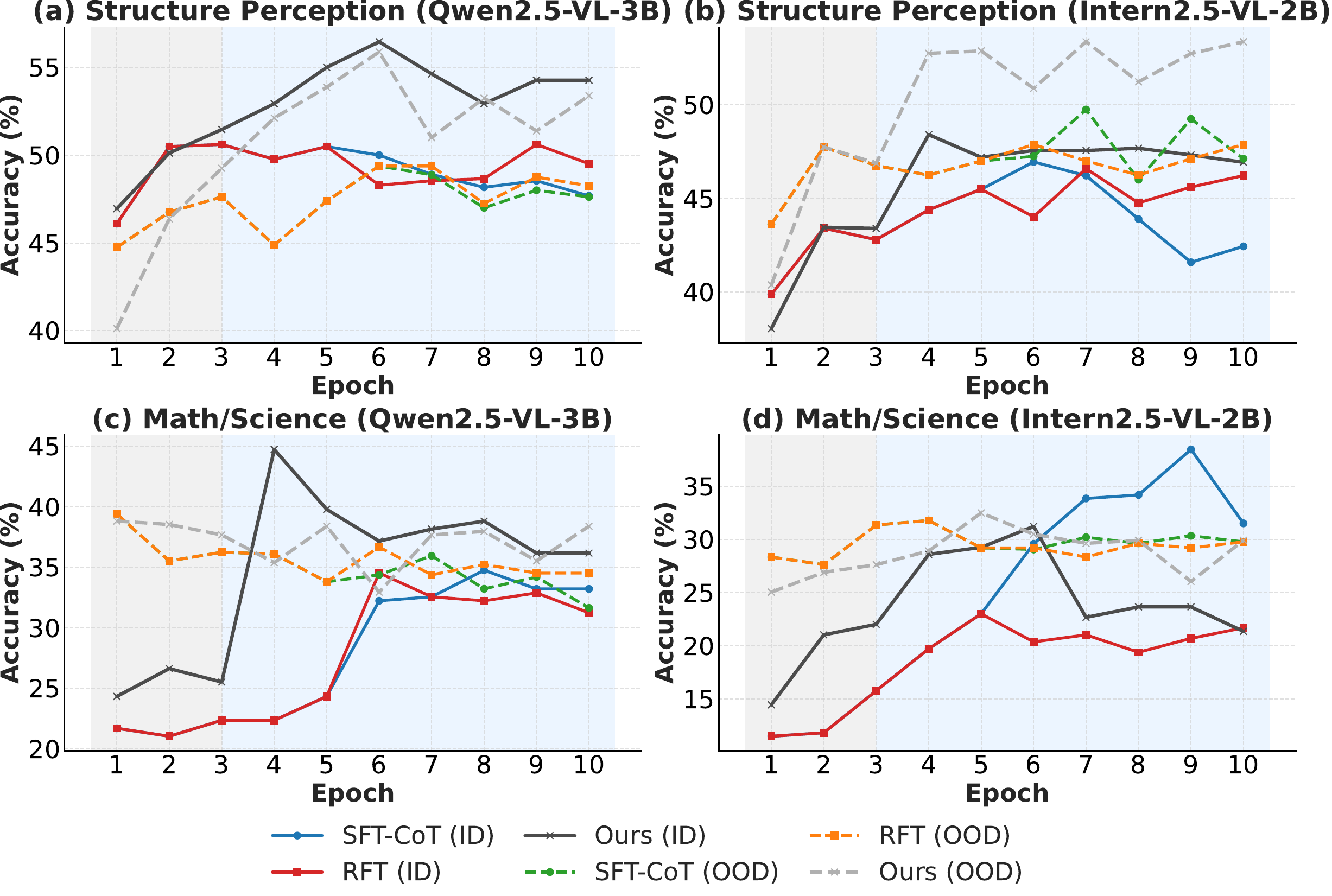}

  \caption{Accuracy over epochs across domains and baselines. Gray region corresponds to the SFT-activation stage, while blue region represent the training stage.}
  \label{fig:acc_steps}
  \vspace{-0.3cm}
\end{figure}


Figure \ref{fig:acc_steps} presents the accuracy trends across epochs under both ID and OOD settings. Overall, our framework mostly outperforms the baseline models throughout training across all domains and model baselines, except InternVL-2B MathVision case, demonstrating improved convergence and stability. Notably, after the first three epochs, a clear performance surge is observed during the training stage, confirming that our framework effectively enhances model optimization beyond the initial SFT-activation phase.






%% file: sec/5_ablation.tex
\subsection{Ablation Study}

\paragraph{Effectiveness of Interleaved Scheme.}
As illustrated in Table \ref{tab:ablation_mcts}, our interleaved GRPO and SFT mostly outperforms both the GRPO only and SFT only variants. As previously discussed, it might be that the GRPO only approach tends to suffer from a cold-start issue, whereas the SFT only variant might prone to error propagation. Furthermore, our method surpasses the variant without log-based reward transformation, indicating that amplifying reward disparities through nonlinear scaling enables GRPO to better distinguish high-quality reasoning paths.
\vspace{1em}

\definecolor{LightBlue}{RGB}{232,243,255}
\begin{table}[t]
\centering
\vspace{-0.2em}
\caption{Ablation results of training variants. GRPO/SFT: Our interleaved GRPO and SFT. GRPO only: Applies GRPO to all reasoning trees without any SFT phase. SFT only: Replaces the GRPO stage with SFT training on the path exhibiting the highest average $\mathcal{Q}(s)$ value for each tree, while keeping the original next SFT phase unchanged. GRPO/SFT w/o trans.: Interleaved GRPO and SFT without log-based nonlinear transformation. }
\resizebox{\columnwidth}{!}{
\begin{tabular}{lcccc}
\toprule
\textbf{Model \& Method} &
\multicolumn{2}{c}{\textbf{Structure Perception}} &
\multicolumn{2}{c}{\textbf{Math/Science}} \\
\cmidrule(lr){2-3} \cmidrule(lr){4-5}
 & GeoMath & Geometry3K & MathVision & DynaMath \\
\midrule
\rowcolor{LightGray}
\multicolumn{5}{l}{\textbf{Qwen2.5-VL-3B}} \\
\midrule
GRPO only & 53.78 & 50.88 & 30.26 & 37.54 \\
SFT only & 45.24 & 35.75 & 27.67 & 37.39  \\
GRPO/SFT \textit{w/o} trans. & 52.32 & 51.5 & 35.86 & \textbf{39.54} \\
\textbf{Ours: GRPO/SFT} & 
\cellcolor{LightBlue}\textbf{55.00} & 
\cellcolor{LightBlue}\textbf{54.37} & 
\cellcolor{LightBlue}\textbf{38.49} & 
\cellcolor{LightBlue} 37.54 \\
\midrule
\rowcolor{LightGray}
\multicolumn{5}{l}{\textbf{Intern2.5-VL-2B}} \\
\midrule
GRPO only & 47.33 & 51.25 & 22.70 & 28.08 \\
SFT only & 47.20 & 51.88 & 25.33 & 29.37 \\
GRPO/SFT \textit{w/o} trans. & 46.46 & 48.62 & 20.39 & 27.65 \\
\textbf{Ours:GRPO/SFT} &
\cellcolor{LightBlue}\textbf{47.68} & 
\cellcolor{LightBlue}\textbf{52.88} & 
\cellcolor{LightBlue}\textbf{31.38} & 
\cellcolor{LightBlue}\textbf{30.66} \\
\bottomrule
\end{tabular}
}
\vspace{-1em}
\label{tab:ablation_mcts}
\end{table}

\definecolor{LightBlue}{RGB}{232,243,255}
\definecolor{LightGray}{RGB}{240,240,240}
\begin{table*}[h]
\centering
\caption{Ablation results of different searching methods during test time. Greedy Search: Expands one node at a time in a step-by-step manner. BestN Search: Expands multiple nodes simultaneously, selecting the top-value node at each step based on PRM-evaluated scores, and continues expanding from the selected node.
MCTS + VLM Judge: Replaces our PRM with a strong VLM (e.g., GPT-4.1~\cite{achiam2023gpt}) to evaluate node values.
MCTS + PRM: Our proposed method.}
\vspace{-0.4em}
\resizebox{0.95\textwidth}{!}{
\begin{tabular}{lccccccc}
\toprule
\textbf{Model \& Method} &
\multicolumn{2}{c}{\textbf{Structure Perception}} &
\multicolumn{3}{c}{\textbf{Spatial Transformation}} &
\multicolumn{2}{c}{\textbf{Math/Science}} \\
\cmidrule(lr){2-3} \cmidrule(lr){4-6} \cmidrule(lr){7-8}
 & GeoMath (ID) & Geometry3K (OOD) & Trance (ID) & TranceL (OOD) & TranceR (OOD) & MathVision (ID) & DynaMath (OOD) \\
\midrule
\rowcolor{LightGray}
\multicolumn{8}{l}{\textbf{Qwen2.5-VL-7B}} \\
\midrule
Greedy Search & 52.32 & 53.37 & \textbf{63.74} & \textbf{52.53} & \textbf{53.20} & 35.53 & 39.68 \\
BestN Search & 55.73 & 52.50 & 63.32 & 51.79 & 52.28 & 44.08 & 42.12 \\
MCTS+VLM Judge	&49.39	&51.25	&60.52	&48.56	&43.12	&36.84	&38.83 \\
\textbf{MCTS+PRM} &
\cellcolor{LightBlue}\textbf{56.10} &
\cellcolor{LightBlue}\textbf{57.38} &
\cellcolor{LightBlue}62.82 &
\cellcolor{LightBlue}50.03 &
\cellcolor{LightBlue}48.60 &
\cellcolor{LightBlue}\textbf{44.74} &
\cellcolor{LightBlue}\textbf{43.55} \\
\bottomrule
\end{tabular}
}
\vspace{-0.5em}
\label{tab:mcts_variants}
\end{table*}

\noindent\textbf{Effectiveness of PRM during Test Time.} Table \ref{tab:mcts_variants} presents the ablation results of different search strategies during test time. Overall, our MCTS + PRM mostly outperforms greedy search, bestN search, and MCTS + VLM Judge across four datasets, demonstrating the effectiveness of our PRM in guiding the search process. The exception occurs on the Trance dataset, where greedy search performs better. This can be attributed to the high lexical and semantic similarity among tokens describing object types, quantities, and positions (e.g., ``circle, left'' vs. ``triangle, right''), which makes fine-grained discrimination more challenging for PRM. Additional model comparisons showing similar trends are provided in Appendix \ref{sec:more_results}.

\begin{figure}[h]
    \centering
    \includegraphics[width=1\linewidth]{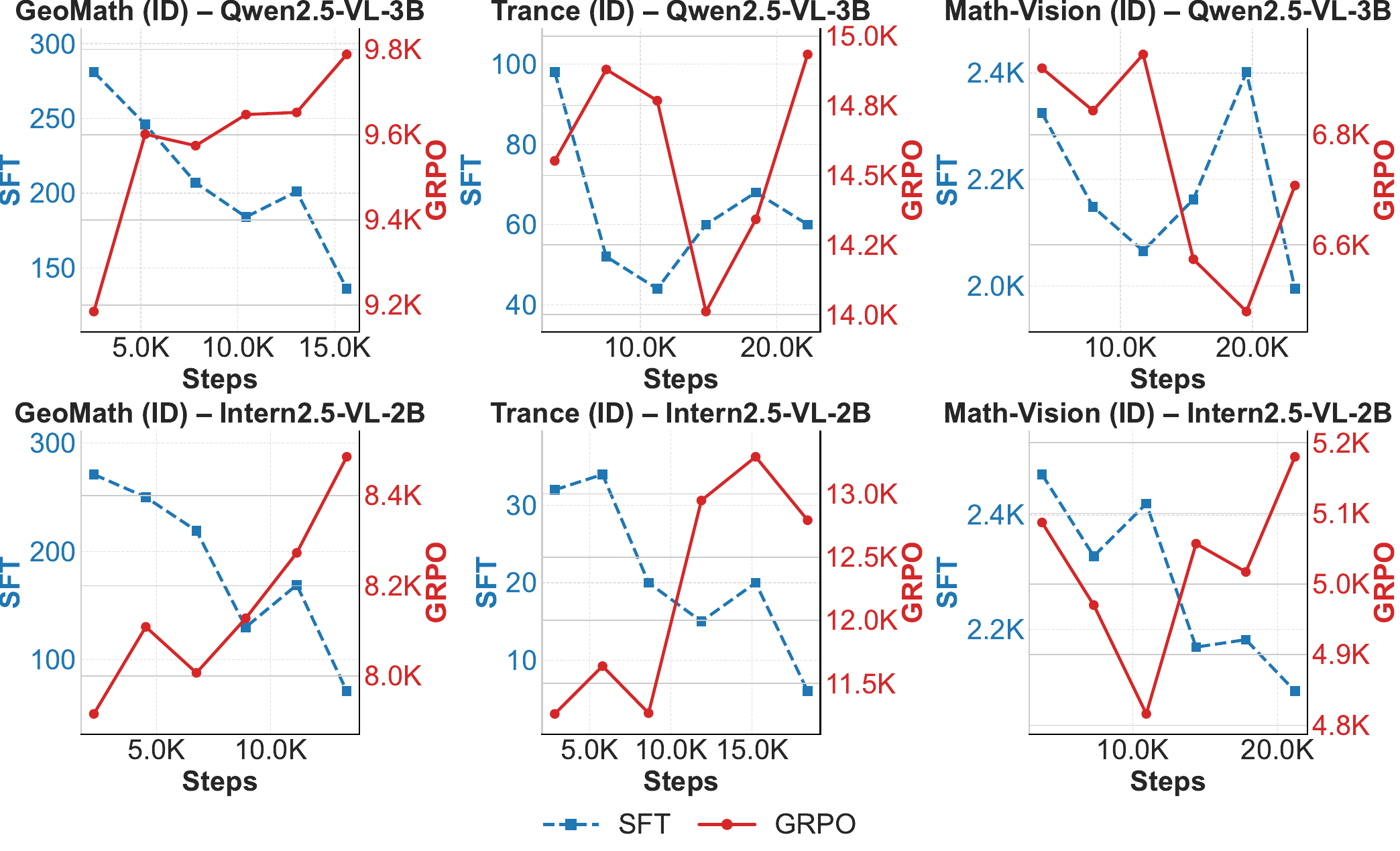}
    \vspace{-1em}
    \caption{Transition of GRPO and SFT data proportions over training steps for Qwen2.5-VL-3B (top row) and Intern2.5-VL-2B (bottom row) across all ID datasets. x-axis is the training steps, y axis left represents number of SFT instances, where y axis right represents number of GRPO instances.}
    \label{fig:training_dynamics_vertical}
    \vspace{-0.5cm}
\end{figure}

\subsection{Transition of Data during Interleaved Training}
Figure \ref{fig:training_dynamics_vertical} illustrates the progressive transition of GRPO and SFT data during training for two models across three ID datasets. As training progresses, we observe that in most cases, the proportion of GRPO-generated data gradually increases, while the reliance on SFT data decreases. This trend aligns with our design expectation: the model initially depends more on supervised fine-tuning to overcome the cold-start stage, and as it becomes more stable and capable, it transitions toward leveraging GRPO to refine and optimize intermediate reasoning steps in greater detail.

%% file: sec/6_case_study.tex
\section{Case Study}

The example below presents a qualitative analysis comparing our framework with three baselines. As shown, the final answer strongly depends on the correctness of intermediate reasoning steps. All the baselines made basic perception error (e.g. recognizing the three interior angles in SFT-CoT output) or logical error (e.g. Given angle is 135 does not imply two angles sums to 45 in RFT output) during the intermediate reasoning process, can propagate and lead to an incorrect final prediction. Our method effectively optimizes these intermediate steps, resulting in a correct final answer. Furthermore, we randomly selected a subset of examples and manually counted the number of perception and logic errors in each example to provide an quantitative analysis. Please refer to the Appendix \ref{sec:case study} for more detailed results, along with additional examples and analyses.

\begin{promptbox}\label{example}
\footnotesize
\textbf{Question:} Determine the value of z. A: 12, B: 13, C: 14, D: 15

\textbf{Ground Truth:} C
\vspace{-0.2em}
\begin{center}
\includegraphics[width=0.5\textwidth]{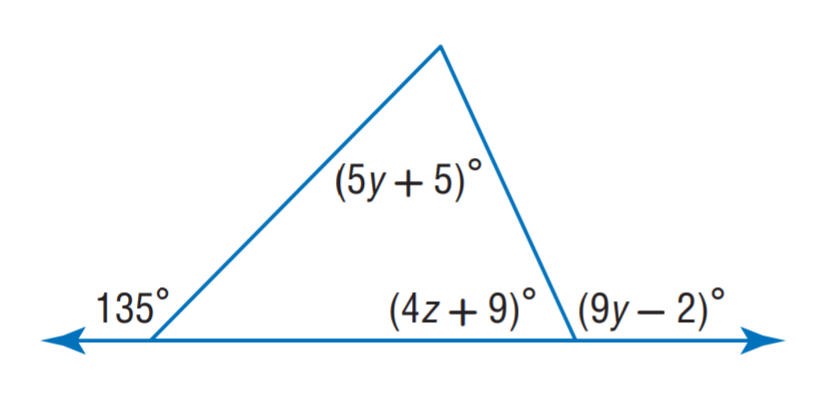}
\end{center}

\vspace{-0.5em}
\textbf{SFT-F Output:} \textcolor{softred}{B}
\vspace{+0.1em}

\textbf{SFT-CoT Output:} Let's start by understanding the problem. We have an exterior angle of 135°, which is equal to the sum of the two interior opposite angles of the triangle. \textcolor{softred}{The interior angles are given as (5y + 5)° and (9y - 2)°}. \textcolor{softred}{We know that the sum of the interior angles of a triangle is 180°. So, (5y + 5) + (4z + 9) + (9y - 2) = 180. ... Simplifying, we get 140 + 4z = 180. ... gives z = 10. The final answer is D}.
\vspace{+0.1em}

\textbf{RFT Output:} Let's analyze the triangle and the straight line formed by its extension. \textcolor{softred}{The sum of angles on a straight line is 180 degrees. The given angle is 135 degrees, we have (4z + 9) = 45 - (9y - 2). ... It seems I made a mistake in my initial assumption. Let's re-evaluate: if z = 12, then 4(12) + 9 = 57, but we need 45 - (9y - 2) = 45 - 17 = 28, which doesn't fit. ...}
\vspace{+0.1em}

\textbf{Our Output:} We have an external angle of 135 degrees, which is equal to the sum of the two opposite internal angles. \textcolor{softgreen}{The internal angles are expressed as (5y + 5) and (4z + 9). First, let's express the sum of the two opposite internal angles: (5y + 5)+ (4z + 9) = 135} Now, let's simplify and solve for z: \textcolor{softgreen}{5y + 4z + 14 = 135, 4z = 135 - 5y - 14, 4z = 121 - 5y}. Since z must be an integer, let's consider the options. \textcolor{softgreen}{Testing z = 14: 4z = 121 - 5y, y = 13. ... Therefore, the correct value of z is 14. The final answer is C}.

\end{promptbox}
\vspace{-0.2em}
\section{Conclusion}
We address two key limitations in current visual reasoning post-training: error propagation and sparse reward signals. To overcome these challenges, we propose PROPA, a unified framework that integrates MCTS with GRPO for dense, process-level optimization. By interleaving SFT and GRPO training and leveraging a PRM to guide test-time search, PROPA effectively aligns inference with the training signal and mitigates reasoning drift. Experiments across seven benchmarks and four VLM backbones demonstrate that PROPA not only improves ID accuracy but also generalizes well to OOD settings, while consistently optimizing the intermediate reasoning process for greater stability.

%% file: sec/7_conclusion.tex

%% file: sec/X_suppl.tex
\clearpage
\setcounter{page}{1}
\maketitlesupplementary

This supplementary material provides additional details and results that complement the main paper. We first present dataset examples and split statistics for the six benchmark datasets used in our evaluations (see \cref{sec:datasets} and Figure~\ref{fig:dataset_examples}), followed by the full set of prompt templates and prompt-design choices employed for activation data creation, SFT, and RFT variants (see \cref{sec:prompts} and Tables~\ref{tab:prompt_types1}--\ref{tab:prompt_types3}). We then report reference training and test runtimes and discuss computational trade-offs introduced by our MCTS-based pre-generation (see \cref{sec:time} and Table~\ref{tab:train_test_time}). Additional ablation results comparing test-time search strategies across multiple VLMs are provided in \cref{sec:more_results} (Table~\ref{tab:mcts_variants_appendix}). Finally, we include qualitative case studies that contrast intermediate-step behavior, perceptual errors, and logic errors between baselines and our method; we also describe a small manual annotation study (n=100) counting perception and logic errors to give an objective error analysis (see \cref{sec:case study} and the example prompt boxes). Together, these materials supply reproducibility details, expanded qualitative evidence, and further empirical analyses supporting the claims in the main paper.

\section{Datasets}
\label{sec:datasets}
\begin{figure*}[t]
\centering
\setlength{\tabcolsep}{4pt}
\renewcommand{\arraystretch}{1.2}

\begin{tabular}{ccc}
\multicolumn{3}{c}{\textbf{(a) Mathematical and Scientific Reasoning}} \\

\begin{subfigure}[t]{0.3\textwidth}
    \includegraphics[width=\linewidth]{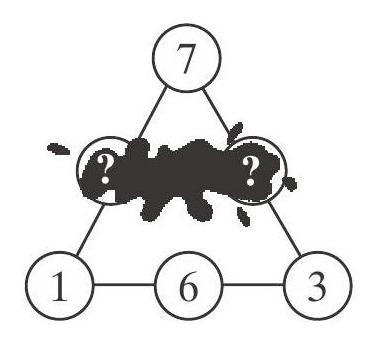}
    \caption*{\textbf{In-domain (MathVision)}\\
    \textit{Q:} The sums of the all the three numbers on each side of the triangle are equal. Two numbers happened to be stained with ink. How much is the sum of these two numbers?\\
    \textit{A:} $2$}
\end{subfigure} &
\begin{subfigure}[t]{0.3\textwidth}
    \includegraphics[width=\linewidth]{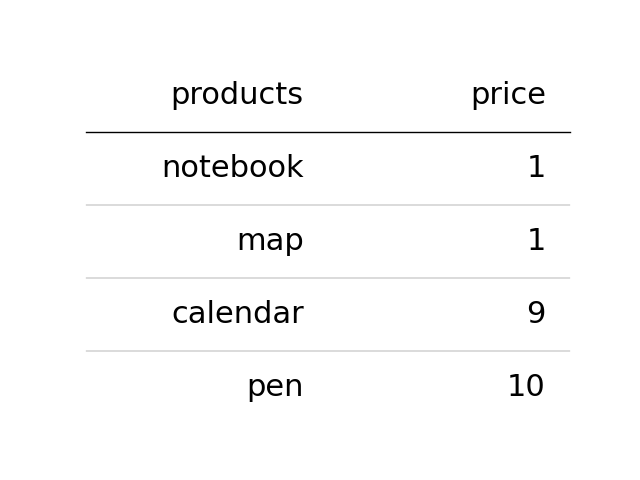}
    \caption*{\textbf{Out-of-domain (DynaMath)}\\
    \textit{Q:} How much money does Hunter need to buy 1 calendars?\\
    \textit{A:} $9$}
\end{subfigure} & \\[4pt]

\multicolumn{3}{c}{\textbf{(b) Spatial Reasoning}} \\

\begin{subfigure}[t]{0.3\textwidth}
    \centering
    \begin{minipage}[t]{0.48\linewidth}
        \includegraphics[width=\linewidth]{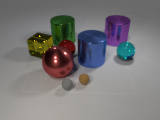}
    \end{minipage}
    \hfill
    \begin{minipage}[t]{0.48\linewidth}
        \includegraphics[width=\linewidth]{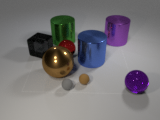}
    \end{minipage}
    \caption*{\textbf{In-domain (Trance)}\\
    \textit{Q:} What transformations are from left to right?\\
    \textit{A:} change\_color(5, brown), change\_position(2, behind), chang\_color(7, gray), change\_color(2, purple)}
\end{subfigure} &
\begin{subfigure}[t]{0.3\textwidth}
    \centering
    \begin{minipage}[t]{0.48\linewidth}
        \includegraphics[width=\linewidth]{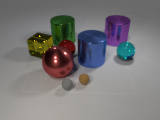}
    \end{minipage}
    \hfill
    \begin{minipage}[t]{0.48\linewidth}
        \includegraphics[width=\linewidth]{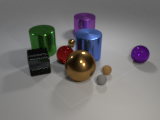}
    \end{minipage}
    \caption*{\textbf{Out-of-domain (TranceL)}\\
    \textit{Q:} What transformations are from left to right?\\
    \textit{A:} change\_color(5, brown), change\_position(2, behind), chang\_color(7, gray), change\_color(2, purple)}
\end{subfigure} &
\begin{subfigure}[t]{0.3\textwidth}
    \centering
    \begin{minipage}[t]{0.48\linewidth}
        \includegraphics[width=\linewidth]{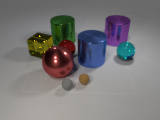}
    \end{minipage}
    \hfill
    \begin{minipage}[t]{0.48\linewidth}
        \includegraphics[width=\linewidth]{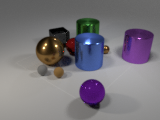}
    \end{minipage}
    \caption*{\textbf{Out-of-domain (TranceR)}\\
    \textit{Q:} What transformations are from left to right?\\
    \textit{A:} change\_color(5, brown), change\_position(2, behind), chang\_color(7, gray), change\_color(2, purple)}
\end{subfigure} \\[4pt]

\multicolumn{3}{c}{\textbf{(c) Structure Perception Reasoning}} \\

\begin{subfigure}[t]{0.3\textwidth}
    \includegraphics[width=\linewidth]{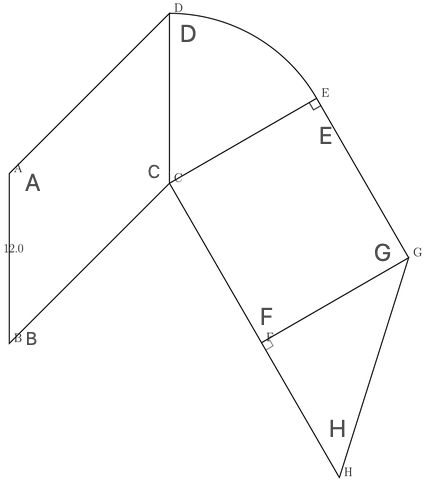}
    \caption*{\textbf{In-domain (GeoMath)}\\
    \textit{Q:} Given that AB measures 12.0 and CDE forms an equilateral triangle, calculate the perimeter of the shape ABCDE. A: 79.5, B: 41.0, C: 56.0, D: 46.5\\
    \textit{A:} C}
\end{subfigure} &
\begin{subfigure}[t]{0.3\textwidth}
    \includegraphics[width=\linewidth]{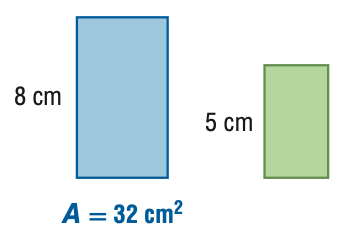}
    \caption*{\textbf{Out-of-domain (Geometry)}\\
    \textit{Q:} For the pair of similar figures, find the area of the green figure. A: 12.5, B: 20.0, C: 51.2, D: 81.9\\
    \textit{A:} A}
\end{subfigure} &
\end{tabular}

\caption{
Example visualization of datasets used in our benchmark. 
We include three reasoning categories: 
(a) Mathematical and Scientific Reasoning, 
(b) Spatial Reasoning, and 
(c) Structure Perception. 
Each category contains both in-domain and out-of-domain datasets, 
highlighting the diversity and reasoning complexity across tasks.
}
\label{fig:dataset_examples}
\end{figure*}

To evaluate our proposed method, we conduct experiments on six benchmark datasets, divided into three in-domain (ID) sets and three corresponding out-of-domain (OOD) sets, covering three distinct and challenging areas of visual reasoning.
All models, including our baselines, are trained and validated on the ID datasets, while final evaluations are performed on the test splits of all six datasets. The first category, \textbf{Mathematical and Scientific Reasoning}, assesses the model’s ability to solve quantitative problems grounded in visual information. For in-domain training, we use MathVision~\cite{wang2024measuring}, a large-scale benchmark of real visual math problems spanning geometry, logic, and science across 16 subjects and five difficulty levels. To test OOD generalization, we use DynaMath~\cite{zou2024dynamath}, a dynamic benchmark constructed from 501 programmatic seed problems involving numerical, geometric, symbolic, and layout transformations. The second category, \textbf{Geometric Structure Understanding}, focuses on analyzing relationships in geometric figures, imaging structures, chart layouts, and architectural designs. The in-domain training data combines Geo170K~\cite{geo170k} and Math360K~\cite{math360k}, offering a broad foundation in geometric reasoning, while the OOD evaluation uses Geometry3K~\cite{lu2021inter}, which features more complex geometric configurations. The third category, \textbf{Visual Spatial Reasoning}, involves spatial-visual reasoning tasks that require models to infer single- or multi-step transformations by analyzing initial and final 3D scenes from multiple viewpoints. Training is performed on Trance~\cite{hong2021transformation}, and generalization to novel perspectives is evaluated using TranceL and TranceR, which present the same scenes from unseen left and right viewpoints, respectively, directly testing the model’s ability to transfer spatial understanding across perspectives. 
Figure \ref{fig:dataset_examples} illustrates representative examples from all seven datasets. As shown, the in-domain and out-of-domain datasets share similar underlying knowledge bases but differ in format or visual representation, allowing us to evaluate the model’s ability to generalize acquired knowledge. In the case of Trance, the out-of-domain variants (Trance-L and Trance-R) are derived from the left and right views of the same set of objects, designed to test the model’s spatial generalizability. We specifically selected these datasets to minimize data contamination issues in current VLMs and ensure that, within each domain, there are at least two variants—one serving as the ID set and another as the OOD set, allowing a controlled evaluation of generalization.
Moreover, Table \ref{tab:dataset_stats} shows the training, valiation, test statistics of all the datasets we used.

\begin{table}[h]
\centering
\caption{
Statistics of dataset splits for each reasoning category. In-domain datasets include train/validation/test splits, while out-of-domain (OOD) datasets contain only test sets.
}
\label{tab:dataset_stats}

\resizebox{\columnwidth}{!}{%
\begin{tabular}{lcccc}
\toprule
\textbf{Category} & \textbf{Dataset} & \textbf{Train} & \textbf{Val} & \textbf{Test} \\
\midrule
\multirow{2}{*}{\textbf{Math}} 
    & MathVision (ID) & 1,000 & 300 & 304 \\
    & DynaMath (OOD) & -- & -- & 698 \\[3pt]
\multirow{3}{*}{\textbf{Spatial}} 
    & Trance (ID) & 1,000 & 300 & 1,000 \\
    & Trance-L (OOD) & -- & -- & 1,000 \\
    & Trance-R (OOD) & -- & -- & 1,000 \\[3pt]
\multirow{2}{*}{\textbf{Structure}} 
    & GeoMath (ID) & 1,000 & 300 & 800 \\
    & Geometry (OOD) & -- & -- & 820 \\
\bottomrule
\end{tabular}%
}
\end{table}

\section{Prompts}
\label{sec:prompts}
\paragraph{Prompt Design.}
Table~\ref{tab:prompt_types1}, Table~\ref{tab:prompt_types2} and Table~\ref{tab:prompt_types3} summarize the prompt types used across different reasoning domains and methods. 
For \textbf{Mathematical and Structural Reasoning} datasets, we employ two types of prompts: 
(1) \textit{Multiple-choice}, where the model selects an answer from several options, and 
(2) \textit{Only-number}, which directly requests a numerical answer.
For \textbf{Spatial Reasoning} datasets, we use a \textit{Spatial} prompt, asking the model to infer spatial relations such as position or orientation.

Each baseline (SFT-F, SFT-CoT, RFT-zero, and RFT), the activation dataset creation and our proposed method use slightly different templates to match their respective reasoning paradigms.

\section{Training and Testing Time}
\label{sec:time}

Table~\ref{tab:train_test_time} reports the reference per epoch training and testing time of all baselines and our method 
on three training datasets (GeoMath, Trance, and Math-Vision), 
evaluated with Qwen2.5-VL-3B using 4×A100 (80GB) GPUs and Intel Xeon Gold 6326 CPU @ 2.90GHz.

Compared with RFT and its variants, our framework introduces a MCTS phase 
to construct an exploration tree before reinforcement optimization. 
During the GRPO stage, we directly utilize the child node values from the MCTS tree to compute the policy loss, 
eliminating the need for the policy model to generate a group of answers. This significantly reduces the computational overhead during reinforcement optimization. However, since our method automatically generates a large number of intermediate states during tree expansion, the overall training time is approximately two times longer in average than baseline RFT-zero and RFT, while inference time shows similar trends.

\begin{table}[h]
\centering
\small
\renewcommand{\arraystretch}{1.2}
\setlength{\tabcolsep}{5pt}
\caption{\textbf{Training and testing time comparison} on three datasets using \textbf{Qwen2.5-VL-3B}. The training time is for 10 epochs and testing time is for whole test set. \textbf{The time is rounded to nearest hour.} Experiments were conducted on 4×A100 (80GB) GPUs and Intel(R) Xeon(R) Gold 6326 CPU @ 2.90GHz.
Our method achieves a more efficient GRPO phase by leveraging MCTS-derived intermediate values, 
though the pre-generation of the exploration tree incurs additional time during training.}
\label{tab:train_test_time}
\resizebox{\columnwidth}{!}{%
\begin{tabular}{lcccccc}
\toprule
\textbf{Dataset} & \textbf{Phase} & \textbf{SFT-F} & \textbf{SFT-CoT} & \textbf{RFT-zero} & \textbf{RFT} & \textbf{Ours} \\
\midrule
\multirow{2}{*}{GeoMath} 
& Training (h) & $<1$ & 1 & 18 & 8 & 38 \\
& Testing (h)  & $<1$  & $<1$  & 1  & 1  & 2 \\
\midrule
\multirow{2}{*}{Trance} 
& Training (h) & $<1$  & 1 & 14 & 12 & 47 \\
& Testing (h)  & $<1$  & $<1$  & 1  & 1  & 2 \\
\midrule
\multirow{2}{*}{MathVision} 
& Training (h) & $<1$  & 2 & 23 & 16 & 29 \\
& Testing (h)  & $<1$  & $<1$  & 1  & 1  & 2 \\
\bottomrule
\end{tabular}
}
\end{table}

\section{More Ablation Results}
\label{sec:more_results}
Table \ref{tab:mcts_variants_appendix} reports the ablation results of different test-time search strategies for three additional VLMs: Intern2.5-VL-2B, Intern2.5-VL-8B, and Qwen2.5-VL-3B. Overall, these models exhibit similar trends to Qwen2.5-VL-7B: our MCTS + PRM consistently outperforms both greedy search and bestN search across four datasets, highlighting the effectiveness of PRM in guiding the search process. However, our search methods are less effective on the Trance dataset.


\section{Case Study}
\label{sec:case study}
The three figures below present three additional examples comparing our framework with other baselines. As shown, baseline models often fail due to perceptual errors in interpreting visual information or logical mistakes in the reasoning process. In contrast, our framework achieves the correct answers by optimizing the intermediate reasoning steps. 

Due to the lack of established metrics for evaluating intermediate reasoning quality, we are unable to provide large-scale quantitative analysis and instead, we randomly selected a subset of examples (n=100) and manually (all co-authors) counted the number of perception and logic errors in each example to provide an more objective analysis. As shown in Table \ref{table:human}, our method consistently yields a significantly lower number of perception and logic errors compared to all other baselines. It can also be seen all models tend to have more perception error than logical error. This qualitative analysis demonstrates that the superior performance of our approach is not solely limited to the final output accuracy, but also reflects a more accurate and reliable intermediate reasoning process.

\begin{table}[h]
\centering
\small
\renewcommand{\arraystretch}{1.2}
\setlength{\tabcolsep}{5pt}
\caption{Human annotation results for the number of Perception Error and Logical Error in GeoMath dataset. The results are reported in the form of ``Number of examples with error / Total Number''. Note that Perception Error and Logical Error might co-exist in the same question.}
\label{table:human}
\resizebox{\columnwidth}{!}{%
\begin{tabular}{lccccc}
\toprule
\textbf{Dataset} & \textbf{Phase}  & \textbf{SFT-CoT} & \textbf{RFT-zero} & \textbf{RFT} & \textbf{Ours} \\
\midrule
\multirow{2}{*}{GeoMath} 
& Perception Error & 31/100 & 38/100 & 36/100 & 23/100 \\
& Logical Error  & 26/100 & 25/100  & 21/100  & 18/100 \\
\bottomrule
\end{tabular}
}
\end{table}

\newtcolorbox{promptbox1}[1][]{
  colback=gray!5!white,
  colframe=black!75!white,
  fonttitle=\bfseries,
  title=MathVision (Qwen2.5-VL-7B),
  #1
}

\begin{promptbox1}\label{example2}
\textbf{Question:} The ladybird want to sit on his flower. The flower has five petals and the stem has three leaves. On which flower should the ladybird sit?

\textbf{Ground Truth:} B
\vspace{-0.2em}
\begin{center}
\includegraphics[width=0.8\textwidth]{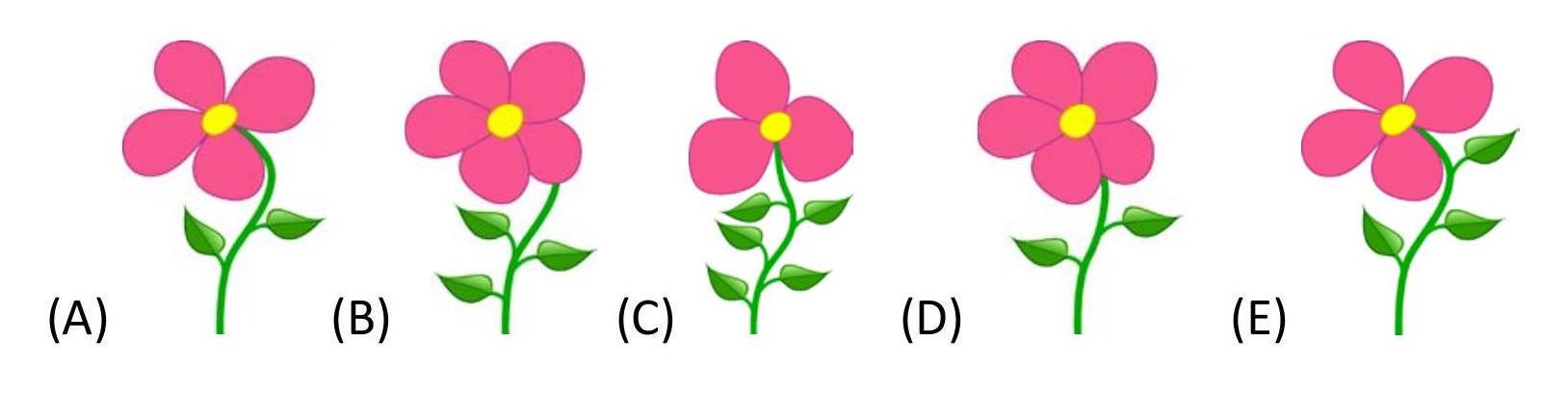}
\end{center}

\vspace{-0.3em}
\textbf{SFT-F Output:} \textcolor{softred}{D}
\vspace{+0.1em}

\textbf{SFT-CoT Output:} Count the number of petals on each flower to find the one with five petals. \textcolor{softred}{Flower E has five petals}. Next, check flower E for having three leaves on its stem. \textcolor{softred}{Flower E's stem has exactly three green leaves}. The final answer is \textcolor{softred}{E}.
\vspace{+0.1em}

\textbf{RFT-zero Output:} Looking at each option: \textcolor{softgreen}{Option A has four petals and two leaves}. Option B \textcolor{softgreen}{has five petals} but \textcolor{softred}{only two leaves}. \textcolor{softred}{Option C has five petals and three leaves}. \textcolor{softred}{Option D has four petals and three leaves}. Option E has \textcolor{softred}{five petals} and \textcolor{softgreen}{three leaves}. \textcolor{softred}{Option C} is the only one that matches both criteria.
\vspace{+0.1em}

\textbf{RFT Output:} Count the number of petals on each flower. \textcolor{softred}{Flowers A, B, C, D, and E all have five petals}. Next, count the number of leaves on the stem for each flower. \textcolor{softgreen}{Flower A has 2 leaves on its stem}, \textcolor{softred}{flower B has 2 leaves}, \textcolor{softred}{flower C has 3 leaves}, \textcolor{softgreen}{flower D has 2 leaves}, and \textcolor{softgreen}{flower E has 3 leaves}. Only \textcolor{softred}{flower E} meets both criteria. \textcolor{softred}{The final answer is E}.
\vspace{+0.1em}

\textbf{Our Output:} Count the number of petals on each flower. \textcolor{softgreen}{Flower A has 4 petals, B has 5 petals}, \textcolor{softred}{C has 4 petals}, \textcolor{softgreen}{D has 5 petals, and E has 4 petals}. Next, count the number of leaves on the stems of each flower. \textcolor{softgreen}{Flower A has 2 leaves, B has 3 leaves}, \textcolor{softred}{C has 3 leaves}, \textcolor{softgreen}{D has 2 leaves, and E has 3 leaves}. Only \textcolor{softgreen}{flower B} matches both criteria. \textcolor{softgreen}{The final answer is B}.

\end{promptbox1}

\newtcolorbox{promptbox2}[1][]{
  colback=gray!5!white,
  colframe=black!75!white,
  fonttitle=\bfseries,
  title=TranceL (Qwen2.5-VL-7B),
  #1
}

\begin{promptbox2}\label{example3}

\textbf{Question:} What is the transformation that can achieve the change from inital state to final state?

\textbf{Ground Truth:} change\_color(5, brown), change\_position(2, behind), change\_color(7, gray), change\_color(2, purple)

\begin{minipage}[t]{0.48\linewidth}
    \includegraphics[width=0.9\linewidth]{images/spatial_out_domain1_1.png}
\end{minipage}
\hfill
\begin{minipage}[t]{0.48\linewidth}
    \includegraphics[width=0.9\linewidth]{images/spatial_out_domain1_2.png}
\end{minipage}

\vspace{-0.3em}
\textbf{SFT-F Output:} \textcolor{softgreen}{change\_color(5, brown)}, \textcolor{softred}{change\_material(7, metal), change\_position(2, behind\_left)}
\vspace{+0.1em}

\textbf{SFT-CoT Output:} We have identified the transformations needed for each object based on the changes observed in the final state compared to the initial state ... \textcolor{softred}{change\_shape(7, cube), change\_size(5, medium)}, \textcolor{softgreen}{change\_color(2, purple)}, \textcolor{softred}{change\_position(6, behind)}
\vspace{+0.1em}

\textbf{RFT-zero Output:} Let's analyze the changes step by step: 1. The large green cylinder remains in the same position and size. 2. The large purple cylinder remains in the same position and size. 3. ... \textcolor{softred}{change\_shape(5, sphere), change\_shape(6, sphere), change\_shape(2, sphere), change\_shape(9, sphere)}
\vspace{+0.1em}

\textbf{RFT Output:} Let's analyze the images and the given attributes to determine the transformations needed. 1. Object 0: Initially a large green metal cylinder. In the final state, it appears to be a medium-sized object. So, we need to change its size. Transformation: `change\_size(0, medium) 2. Object 1: ... \textcolor{softred}{change\_size(0, medium), change\_position(1, front\_right), change\_color(2, gray), change\_shape(3, sphere), change\_size(5, medium)}, \textcolor{softgreen}{change\_color(7, gray)}
\vspace{+0.1em}

\textbf{Our Output:} First, let's focus on object 5, which is initially a large red metal sphere. In the final state, we see a large brown sphere. \textcolor{softgreen}{This indicates that object 5 has undergone a color change from red to brown}. So, the first transformation is \textcolor{softgreen}{change\_color(5, brown)}. Next, let's look at object 7, ... \textcolor{softgreen}{has undergone a color change from yellow to gray}. So, the second transformation is \textcolor{softgreen}{change\_color(7, gray)}. Finally, ... By applying these transformations, we can achieve the change from the initial state to the final state. The final answer is \textcolor{softgreen}{change\_color(5, brown), change\_color(7, gray), change\_color(2, purple)}, \textcolor{softred}{missing change\_position(2, behind)}

\end{promptbox2}

\newtcolorbox{promptbox3}[1][]{
  colback=gray!5!white,
  colframe=black!75!white,
  fonttitle=\bfseries,
  title=DynaMath (Qwen2.5-VL-7B),
  #1
}

\begin{promptbox3}\label{example4}
\textbf{Question:} Find the value of the last row of calculations.

\textbf{Ground Truth:} 12

\begin{center}
\includegraphics[width=0.6\textwidth]{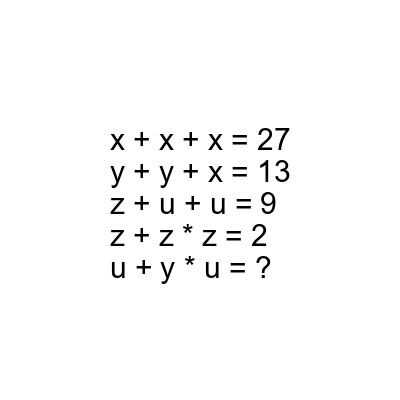}
\end{center}

\vspace{-0.3em}
\textbf{SFT-F Output:} \textcolor{softred}{10}
\vspace{+0.1em}

\textbf{SFT-CoT Output:} x = 9, y = 2, \textcolor{softred}{u = 0}, \textcolor{softred}{z = 0}, so u + y * u = 0 + 2 * 0 = 0. The final answer is \textcolor{softred}{0}.
\vspace{+0.1em}

\textbf{RFT-zero Output:} \textcolor{softgreen}{Starting with x + x + x = 27, we find that x = 9}. \textcolor{softgreen}{Then y + y + x = 13 becomes y + y + 9 = 13, so y = 2}. Next, z + u + u = 9 simplifies to z + 2u = 9. \textcolor{softred}{And z + z * z = 2 can be rewritten as z * z + z = 2, which gives z = -2 (since z cannot be positive for this equation)}. \textcolor{softred}{Substituting z = -2 into z + 2u = 9 gives -2 + 2u = 9, so u = 5.5}. Finally, \textcolor{softred}{u + y * u = 5.5 + 2 * 5.5 = 5.5 + 11 = 16.5.}
\vspace{+0.1em}

\textbf{RFT Output:} \textcolor{softgreen}{x = 9 because 27 ÷ 3 = 9}. \textcolor{softred}{y = 5 because 13 ÷ 2 = 6.5}, but let's try integer solutions: 13 = 2y + x, x = 9, y = 2. \textcolor{softgreen}{z = 1 because 2 = z + z * z, so z = 1}. \textcolor{softred}{u = 0 because 9 = z + u + u}, and \textcolor{softgreen}{z = 1, so 1 + u + u = 9, 2u = 8, u = 4}. Substitute values into the last expression: \textcolor{softgreen}{u + y × u = 4 + 2 × 4 = 4 + 8 = 12}. The final answer is \textcolor{softgreen}{12}.

\textbf{Our Output:} \textcolor{softgreen}{From x + x + x = 27, we get 3x = 27, so x = 9. From y + y + x = 13 and knowing x = 9, we have 2y + 9 = 13, so 2y = 4 and y = 2}. \textcolor{softgreen}{We have z = 1 from 2 = z + z * z. From z + u + u = 9, substitute u = 4, giving 1 + 4 + 4 = 9}. Now calculate the last row: \textcolor{softgreen}{u + y * u = 4 + 2 * 4 = 4 + 8 = 12. The final answer is 12.}

\end{promptbox3}

\begin{table*}[h]
\centering
\small
\renewcommand{\arraystretch}{1.2}
\setlength{\tabcolsep}{5pt}
\caption{\textbf{Prompt types for Activation dataset creation, SFT-F, and SFT-CoT used across different datasets and methods.} 
Multiple-choice and Only-number prompts are used for Mathematical and Structural reasoning datasets, 
while Spatial prompts are used for Spatial reasoning datasets.}
\label{tab:prompt_types1}
\begin{tabular}{{p{4cm} p{3cm} p{9cm}}}
\toprule
\textbf{Prompt Type} & \textbf{Applicable Datasets} & \textbf{Prompt} \\
\midrule
Activation Dataset Creation & All & Think step by step to solve the problem in the image: \texttt{original\_question}. The ground truth final answer is \texttt{final\_answer}, make sure the derived final answer is the same as the ground truth final answer (i.e. Ending with 'The final answer is \texttt{final\_answer}.') and do NOT directly use the correct final answer to derive the steps. Do not describe what should be done. Directly perform the reasoning or calculation at each step. At each step, give only the actual action/result in one concise sentence. Avoid any general explanation, plan, or meta-reasoning. \\
Multiple-choice (SFT-F) & GeoMath, Geometry, MathVision, DynaMath & You are a visual reasoning expert. \texttt{original\_question} Choose one option from the following: \texttt{options}. Please ONLY respond with the selected option letter.  \\
Number Only (SFT-F) & GeoMath, Geometry, MathVision, DynaMath & You are a visual reasoning expert. \texttt{original\_question} Please ONLY respond with a number.  \\
Structured Answer (SFT-F) & Trance, TranceL, TranceR & You are a visual reasoning expert. The first image shows the central view of initial state, the second image shows the final state, and the attributes of the initial objects are listed below: (0, cube, small, gray, glass) (1, sphere, medium, yellow, glass) ... What is the transformation process that can achieve the change from inital state to final state? Describe the changes using only the following format: 'change\_xxx(object\_index, attribute\_after\_change)' Here, xxx can be shape, size, position, color, or material. object\_index can be object's index number, and attribute\_after\_change is the new value of the attribute. When there are multiple changes, list them in a single line, separated by a comma and a space.  \\

Multiple-choice (SFT-CoT) & GeoMath, Geometry, MathVision, DynaMath & You are a visual reasoning expert. \texttt{original\_question} Choose one option from the following: \texttt{options}. The final answer should be a option letter. Please ONLY response one step at a time. Think step by step to solve this problem. Respond with reasoning in \texttt{<think></think>} and only the option in \texttt{<answer></answer>}.  \\
Number Only (SFT-CoT) & GeoMath, Geometry, MathVision, DynaMath & You are a visual reasoning expert. \texttt{original\_question} Please provide a step-by-step reasoning process with calculations to arrive at the final area. The final answer should be a number. Please ONLY response one step at a time. Think step by step to solve this problem. Respond with reasoning in \texttt{<think></think>} and only the number in \texttt{<answer></answer>}.  \\
Structured Answer (SFT-CoT) & Trance, TranceL, TranceR & You are a visual reasoning expert. The first image shows the central view of initial state, the second image shows the final state, and the attributes of the initial objects are listed below: (0, cube, small, gray, glass) (1, sphere, medium, yellow, glass) ... What is the transformation process that can achieve the change from inital state to final state? Describe the changes using only the following format: 'change\_xxx(object\_index, attribute\_after\_change)' Here, xxx can be shape, size, position, color, or material. object\_index can be object's index number, and attribute\_after\_change is the new value of the attribute. When there are multiple changes, list them in a single line, separated by a comma and a space. Think step by step to solve this problem. Respond with reasoning in \texttt{<think></think>} and only the final answer in \texttt{<answer></answer>}. \\

\bottomrule
\end{tabular}
\end{table*}

\begin{table*}[h]
\centering
\small
\renewcommand{\arraystretch}{1.2}
\setlength{\tabcolsep}{5pt}
\caption{\textbf{Prompt types for RFT-zero and RFT used across different datasets and methods.} 
Multiple-choice and Only-number prompts are used for Mathematical and Structural reasoning datasets, 
while Spatial prompts are used for Spatial reasoning datasets.}
\label{tab:prompt_types2}
\begin{tabular}{{p{4cm} p{3cm} p{9cm}}}
\toprule
\textbf{Prompt Type} & \textbf{Applicable Datasets} & \textbf{Prompt} \\
\midrule
Multiple-choice (RFT-zero and RFT) & GeoMath, Geometry, MathVision, DynaMath & You are a visual reasoning expert. \texttt{original\_question} Respond with short and concise reasoning steps in \texttt{<think></think>} and ONLY a final answer number in \texttt{<answer></answer>}.  \\
Number Only (RFT-zero and RFT) & GeoMath, Geometry, MathVision, DynaMath & You are a visual reasoning expert. \texttt{original\_question} Choose one option from the following: \texttt{options}. Respond with short and concise reasoning steps in \texttt{<think></think>} and ONLY a final answer option letter in \texttt{<answer></answer>}.  \\
Structured Answer (RFT-zero and RFT) & Trance, TranceL, TranceR & You are a visual reasoning expert. The first image shows the central view of initial state, the second image shows the final state, and the attributes of the initial objects are listed below: (0, cube, small, gray, glass) (1, sphere, medium, yellow, glass) ... What is the transformation process that can achieve the change from inital state to final state? Describe the changes using only the following format: 'change\_xxx(object\_index, attribute\_after\_change)' Here, xxx can be shape, size, position, color, or material. object\_index can be object's index number, and attribute\_after\_change is the new value of the attribute. When there are multiple changes, list them in a single line, separated by a comma and a space. Think step by step to solve this problem. Respond with reasoning in \texttt{<think></think>} and only the final answer in \texttt{<answer></answer>}. \\
\bottomrule
\end{tabular}
\end{table*}

\begin{table*}[h]
\centering
\small
\renewcommand{\arraystretch}{1.2}
\setlength{\tabcolsep}{5pt}
\caption{\textbf{Prompt types for Our methods used across different datasets and methods.} 
Multiple-choice and Only-number prompts are used for Mathematical and Structural reasoning datasets, 
while Spatial prompts are used for Spatial reasoning datasets.}
\label{tab:prompt_types3}
\begin{tabular}{{p{4cm} p{3cm} p{9cm}}}
\toprule
\textbf{Prompt Type} & \textbf{Applicable Datasets} & \textbf{Prompt} \\
\midrule
Multiple-choice (Ours) & GeoMath, Geometry, MathVision, DynaMath & \textit{System Prompt:} A conversation between User and Assistant. The input by user may include some existing steps to solve the question and the Assistant should continue to derive only the next step based on these existing steps. If the input does not provide any existing steps, the Assistant need to analyze the problem and then give the first step in solving the problem. If the Assistant think it has reached the final step, provide the final answer choice following the format 'The final answer is ...', otherwise continue to the next inference step. \textit{User Prompt:} Think step by step to solve the problem in the image: \texttt{original\_question}. The final answer should be a option letter. Please ONLY response one step at a time. Existing Steps: \texttt{previous\_steps} Next Step: \\
Number Only (Ours) & GeoMath, Geometry, MathVision, DynaMath & \textit{System Prompt:} A conversation between User and Assistant. The input by user may include some existing steps to solve the question and the Assistant should continue to derive only the next step based on these existing steps. If the input does not provide any existing steps, the Assistant need to analyze the problem and then give the first step in solving the problem. If the Assistant think it has reached the final step, provide the final answer number following the format 'The final answer is ...', otherwise continue to the next inference step. \textit{User Prompt:} Think step by step to solve the problem in the image: \texttt{original\_question}. The final answer should be a number. Please ONLY response one step at a time. Existing Steps: \texttt{previous\_steps} Next Step: \\
Structured Answer (Ours) & Trance, TranceL, TranceR & \textit{System Prompt:} A conversation between User and Assistant. The input by user may include some existing steps to solve the question and the Assistant should continue to derive only the next step based on these existing steps. If the input does not provide any existing steps, the Assistant need to analyze the problem and then give the first step in solving the problem. If the Assistant think it has reached the final step, provide the final answer following the format 'The final answer is ...', otherwise continue to the next inference step. \textit{User Prompt:} Think step by step to solve the problem in the image: \texttt{original\_question}. Please ONLY response one step at a time. Existing Steps: \texttt{previous\_steps} Next Step: \\

\bottomrule
\end{tabular}
\end{table*}

\definecolor{LightBlue}{RGB}{232,243,255}
\begin{table*}[h]
\centering
\caption{Ablation results of different searching methods during test time. Greedy search: Expands one node at a time in a step-by-step manner. BestN search: Expands multiple nodes simultaneously, selecting the top-value node at each step based on PRM-evaluated scores, and continues expanding from the selected node. MCTS + PRM: Our proposed method.}
\resizebox{\textwidth}{!}{
\begin{tabular}{lccccccc}
\toprule
\textbf{Model \& Method} &
\multicolumn{2}{c}{\textbf{Structure Perception}} &
\multicolumn{3}{c}{\textbf{Spatial Transformation}} &
\multicolumn{2}{c}{\textbf{Math/Science}} \\
\cmidrule(lr){2-3} \cmidrule(lr){4-6} \cmidrule(lr){7-8}
 & GeoMath (ID) & Geometry3K (OOD) & Trance (ID) & TranceL (OOD) & TranceR (OOD) & Math-Vision (ID) & DynaMath (OOD) \\
\midrule
\multicolumn{8}{l}{\textbf{Qwen2.5-VL-3B}} \\
\midrule
greedy search & 51.22 & 46.75 & 54.13 & 46.64 & \textbf{48.02} & 34.21 & 39.54 \\
bestN search & 54.15 & 51.12 & \textbf{57.65} & \textbf{48.75} & 47.81 & 34.54 & 36.10 \\
\textbf{MCTS+RPM} &
\cellcolor{LightBlue}\textbf{55.00} &
\cellcolor{LightBlue}\textbf{54.37} &
\cellcolor{LightBlue}56.32 &
\cellcolor{LightBlue}47.59 &
\cellcolor{LightBlue}45.70 &
\cellcolor{LightBlue}\textbf{38.49} &
\cellcolor{LightBlue}\textbf{37.54} \\
\midrule
\multicolumn{8}{l}{\textbf{Intern2.5-VL-2B}} \\
\midrule
greedy search & 46.71 & 47.25 & \textbf{63.54} & \textbf{54.12} & \textbf{53.04} & 23.68 & 26.50 \\
bestN search & 46.46 & 48.00 & 60.31 & 48.53 & 49.62 & 20.39 & 27.51 \\
\textbf{MCTS+PRM} &
\cellcolor{LightBlue}\textbf{47.68} &
\cellcolor{LightBlue}\textbf{52.88} &
\cellcolor{LightBlue} 59.13 &
\cellcolor{LightBlue} 47.37 &
\cellcolor{LightBlue} 46.69 &
\cellcolor{LightBlue}\textbf{31.38} &
\cellcolor{LightBlue}\textbf{30.66} \\
\midrule
\multicolumn{8}{l}{\textbf{Intern2.5-VL-8B}} \\
\midrule
greedy search & 52.80 & 58.13 &  \textbf{77.62} & 62.08 & \textbf{62.82} & 34.54 & 34.96 \\
bestN search & 57.68 & 62.50 & 75.32 & 60.78 & 60.82 & 31.25 & 34.10 \\
\textbf{MCTS+PRM} &
\cellcolor{LightBlue}\textbf{58.41} &
\cellcolor{LightBlue}\textbf{61.00} &
\cellcolor{LightBlue} 74.33 &
\cellcolor{LightBlue} \textbf{62.63} &
\cellcolor{LightBlue} 61.69 &
\cellcolor{LightBlue}\textbf{38.16} &
\cellcolor{LightBlue}\textbf{40.10} \\
\bottomrule
\end{tabular}
}
\vspace{-2mm}
\label{tab:mcts_variants_appendix}
\end{table*}